\documentclass[10pt,twocolumn,letterpaper]{article}

\usepackage[pagenumbers]{cvpr} 

\definecolor{cvprblue}{rgb}{0.21,0.49,0.74}
\usepackage[pagebackref,breaklinks,colorlinks,allcolors=cvprblue]{hyperref}
\usepackage{xcolor}
\usepackage{multirow}
\usepackage{colortbl}

\def\paperID{} 
\def\confName{CVPR}
\def\confYear{2026}

\usepackage{stfloats}

\title{Bridging Video Understanding and Generation in a Unified Framework}
\newcommand{\methodname}{Vega}
\author{
 Yuqi~Wang$^{*}$ \quad Runyi~Li$^{*}$ \quad Ruoyu~Feng \quad Renjie~Chen \quad Wenfeng~Lin \quad Mingyu~Guo \\[1mm]
}

\begin{document}
\twocolumn[{%
\renewcommand\twocolumn[1][]{#1}%
\maketitle
\begin{center}
    \centering
    \includegraphics[width=\textwidth]{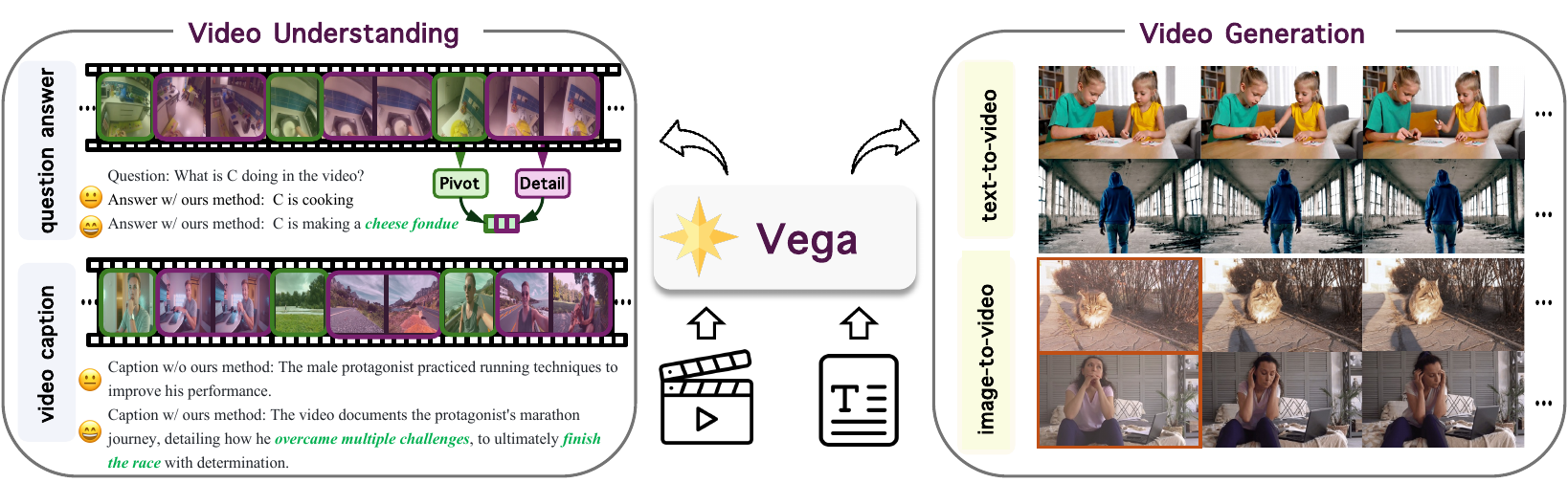} 
    \vspace{-4mm}
    \captionof{figure}{\textbf{Overview of \methodname{}, a unified framework that bridges video generation and understanding.} Our model supports both video and text inputs, enabling video understanding tasks such as video question answering and captioning, as well as video generation tasks including text-to-video and image-to-video generation within a single framework.}
    \label{fig:teaser}
    \vspace{-1mm}
\end{center}}]

\begin{abstract}
Recently, unified image generation and understanding have been extensively explored. However, extending such unified modeling paradigms to the video domain remains largely underexplored. A central challenge is that video understanding favors compact, discriminative semantic representations, whereas video generation requires dense signals that preserve visual details and temporal coherence. Videos naturally capture both spatial semantics and temporal dynamics, making them a more suitable modality for unified multimodal modeling compared to static images. In this paper, we propose Vega, a unified framework that bridges video understanding and generation. Vega leverages a shared vocabulary to jointly model text and visual representations and employs a hybrid architecture combining autoregressive (AR) prediction with diffusion-based rendering. Specifically, the AR model focuses on predicting semantically meaningful visual tokens for keyframes, providing a structured representation that guides the diffusion module in rendering dense, high-resolution video frames. Extensive experiments demonstrate that Vega achieves strong performance on video generation benchmarks such as VBench and video understanding benchmarks like VideoMME.
\end{abstract}
    
\section{Introduction}
\label{sec:intro}
Recent years have witnessed the remarkable success of unified models in natural language processing, evolving from task-specific models to Large Language Models (LLMs) as unified foundation models.
Inspired by this paradigm, researchers have also sought to develop unified models in the vision domain~\cite{wang2023images,bai2024sequential,ravishankar2025scaling}.
In the image domain, unified models for generation and understanding have attracted significant attention~\cite{xie2024towards, zhang2025unified}, leading to the exploration of diverse architectures, including purely autoregressive (AR)~\cite{team2024chameleon, wu2024vila, wang2024emu3}, hybrid AR–diffusion~\cite{zhou2024transfusion,chen2025blip3, ma2025janusflow, deng2025emerging, wu2025omnigen2, liao2025mogao, xie2025show, geng2025x}, and fully diffusion~\cite{yang2025mmada, swerdlow2025unified}, all aimed at learning a single model with both understanding and generative capabilities.
However, current research efforts remain largely confined to static images, with limited exploration of unified modeling for videos.

The success of LLMs reflects a profound idea: in language, \emph{to generate is to understand}. Extending this notion to vision, video emerges as a more natural form for unified modeling, embodying both spatial semantics and the temporal–causal structure of the real world. 
Recent studies~\cite{wiedemer2025video, guo2025video, tong2025thinking} have further echoed this philosophy: by generating what it understands, and understanding what it generates, video models—such as Sora2 and Veo3—demonstrate that video generation can act as a universal interface for visual understanding, enabling zero-shot reasoning and segmentation across diverse tasks.
Despite this progress, existing research on video understanding and video generation remains largely disjoint. The former is primarily driven by vision–language models (VLMs)~\cite{cheng2024videollama,  wang2025internvideo2, bai2025qwen2} that emphasize visual recognition and reasoning, whereas the latter is advanced by diffusion-based generative paradigms~\cite{blattmann2023stable, yang2024cogvideox, kling16,kong2024hunyuanvideo,wan2025wan}, which focus on producing visually realistic and temporally coherent videos. This disconnect motivates a fundamental question: \emph{can we bridge these two paradigms within a single, unified framework?}

Bridging these paradigms is inherently challenging, as video understanding and generation demand fundamentally different types of representations. 
Understanding tasks require compact, discriminative semantic embeddings for recognition and reasoning, while generation necessitates dense, high-fidelity signals to synthesize realistic and temporally coherent visual content. Prior efforts can be broadly categorized into two directions.
The first investigates discrete, fully autoregressive (AR) formulations~\cite{wu2024vila,wang2024emu3}, extending image-based modeling to the video domain. However, these extensions often underperform—autoregressive video generation remains substantially inferior to diffusion-based methods in terms of visual fidelity and temporal consistency. 
This limitation arises from the intrinsically low signal-to-noise ratio (SNR) in video data, which causes pixel-space VQ compression to generate an excessive number of visual tokens and introduces strong temporal redundancy that hinders effective learning.
The second line of work explores hybrid AR–diffusion architectures~\cite{luo2025univid, wei2025univideo}, which integrate multimodal large language models (MLLMs) with diffusion decoders, serving as replacements for traditional text encoders such as T5~\cite{raffel2020exploring}.
While such hybrid designs enhance in-context generation quality, they inherently prioritize generative capabilities. Their understanding components still depend on frozen MLLMs, limiting the potential to achieve a truly unified framework that bridges visual understanding and generation.
Despite these efforts, existing approaches either suffer from inferior visual quality or depend on disjoint modules for generation and understanding respectively, underscoring the need for a unified solution that can bridge both paradigms.

To address these challenges, we introduce \methodname{}, a unified framework for video generation and understanding.
Our approach harnesses the complementary strengths of autoregressive (AR) and diffusion modeling: the AR module operates in a semantic space, predicting keyframe-level representations that capture high-level scene dynamics, while the diffusion decoder serves purely as a pixel-level renderer, translating these semantic embeddings into continuous and realistic video sequences.
At the core of \methodname{} lies the use of semantic vision tokens as a shared representation, bridging the gap between understanding and generation. This not only strengthens the semantic alignment with text tokens but also significantly reduces the number of visual tokens.
As for generation tasks, we adopt a unified video-centric modeling approach that encompasses a variety of tasks. In this formulation, image editing can be treated as a special case of a two-frame video~\cite{rotstein2025pathways}. To mitigate the loss of fine-grained details by semantic tokens and ensure temporal consistency, we employ noise-controlled visual conditioning, which enforces coherence in the generated visual content. On the understanding side, motivated by the slowfast concept~\cite{feichtenhofer2019slowfast}, we introduce a dual-flow selection mechanism to further compress the number of visual tokens, and incorporate a generative visual loss as auxiliary supervision. Our experiments suggest that this auxiliary supervision can provide additional learning signals for video understanding, highlighting the potential benefits of integrating generation and understanding within a single framework.

Our contributions can be summarized as follows:
\begin{itemize}
    \item We propose \textbf{\methodname{}, a unified framework that bridges video generation and understanding} through a shared representation space. By converting both textual and visual signals into a unified set of discrete tokens, \methodname{} enables a consistent multimodal representation that supports both generative and understanding tasks.
    \item Unlike prior autoregressive generation methods that directly model pixel-level dependencies, our framework performs autoregression in the \textbf{visual semantic space}. The AR module models high-level scene dynamics by predicting semantic representations for keyframes, which the diffusion decoder refines into high-fidelity videos. This design establishes a complementary interplay: the AR module learns state transitions in the latent semantic space, while the diffusion decoder renders detailed appearances in pixel space.
    \item We propose a \textbf{dual-flow selection} mechanism for visual tokens in video understanding, comprising complementary \textit{pivot} and \textit{detail} flows that capture temporal dynamics across multiple timescales. The detail flow focuses on rapidly changing features under higher spatial compression, while the pivot flow models slowly varying patterns with lower compression. This design enables efficient hierarchical temporal modeling and significantly reduces the number of visual tokens. Furthermore, incorporating \textbf{generative supervision} on vision tokens provides an auxiliary signal for video understanding.
    \item \methodname{} achieves leading performance among unified video understanding-generation models, while showing competitive results against specialized models on both video generation and video understanding benchmarks. Moreover, it unifies image-to-video (I2V) generation and image editing under a single video-centric modeling framework.
\end{itemize}

\section{Related Works}

\subsection{Unified Multimodal Understanding and Generation}
Previous VLMs based on autoregressive (AR) modeling and generative models based on diffusion modeling have achieved remarkable success in visual understanding and generation, respectively. Recently, unified multimodal understanding and generation~\cite{zhang2025unified} has attracted growing attention, leading to a series of approaches for the image modality, including pure AR~\cite{team2024chameleon, wang2024emu3, xie2025muse}, hybrid AR-diffusion~\cite{dong2023dreamllm, zhou2024transfusion, chen2025blip3, ma2025janusflow, chen2025janus, deng2025emerging, liao2025mogao, lin2025uniworld, chen2025blip3o}, and pure diffusion paradigms~\cite{yang2025mmada, swerdlow2025unified}.
Purely autoregressive approaches, such as Chameleon~\cite{team2024chameleon} and Emu3~\cite{wang2024emu3}, represent images and videos as discrete tokens using VQ-VAE~\cite{van2017neural}, and train models in an autoregressive manner.
Hybrid autoregressive–diffusion models, exemplified by Janus-Pro~\cite{chen2025janus} and Bagel~\cite{deng2025emerging}, enable generation and understanding to proceed through distinct pathways, while supporting joint training within a unified framework. Pure diffusion-based approaches constitute another emerging research direction. Representative works such as MMaDA~\cite{yang2025mmada} leverage diffusion models to achieve unified modeling of images and text.

Extending these efforts beyond the image modality, recent studies~\cite{tan2025omni, luo2025univid, xie2025show, wei2025univideo} have begun to explore unified modeling for videos. However, most existing approaches condition diffusion models solely on text embeddings enhanced by MLLMs, focusing primarily on improving the expressiveness of textual representations.
In contrast, our framework emphasizes the joint learning of vision–language representations within the MLLM itself, achieving a tighter integration between the two modalities.
Specifically, the autoregressive branch predicts the visual tokens of key frames, which are then decoded by the diffusion module into temporally dense video sequences.

\subsection{Autoregressive Video Generation}
Despite the success of diffusion models in video generation, recent studies have increasingly explored autoregressive alternatives~\cite{wang2024emu3, wu2024vila, deng2024autoregressive}. Existing approaches can be broadly grouped into two main paradigms.

\noindent{\textbf{Discrete token-based models}}, which leverage discrete VQ tokens and autoregressive transformers for video generation (e.g., VideoPoet~\cite{kondratyuk2023videopoet}, Emu3~\cite{wang2024emu3}, Lonng~\cite{wang2024loong}, VILA-U~\cite{wu2024vila}, COSMOS~\cite{agarwal2025cosmos}). 
Other works, such as Lumos-1~\cite{yuan2025lumos} and Show-o~\cite{xie2024show}, also adopt discrete tokens, but model them through discrete diffusion—often implemented as masked autoregressive denoising—rather than conventional next-token prediction.

\noindent{\textbf{Continuous latent-based models}}, which operate on VAE latents and receive supervision through diffusion heads, draw inspiration from the recent success of diffusion models~\cite{rombach2022high, blattmann2023stable}.
For instance, NOVA~\cite{deng2024autoregressive} and VideoMAR~\cite{yu2025videomar} extend MAR~\cite{li2024autoregressive} to the video domain by employing a diffusion head to autoregressively model spatio-temporal latent sequences.

However, prior approaches—whether based on discrete tokens or continuous latents—primarily model videos in the pixel space.
In contrast, our method performs autoregressive video prediction in the semantic space~\cite{tschannen2025siglip}.

\begin{figure*}[ht]
  \centering
    \includegraphics[width=\textwidth]{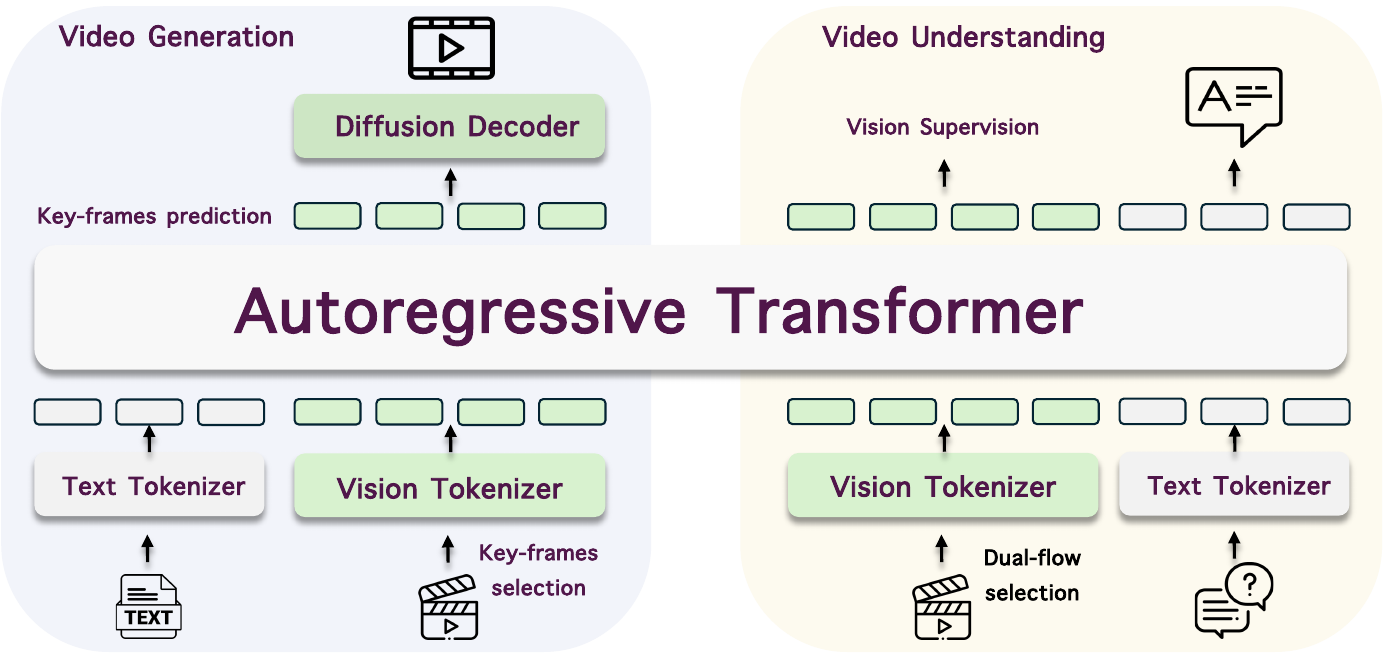} 
  \caption{\textbf{Illustration of our \methodname{} framework.} Both textual and visual inputs are converted into discrete tokens and trained jointly within an autoregressive transformer, where vision and language share a unified codebook. For video generation, the model uniformly samples video frames and predicts keyframe-level visual representations, which are further rendered into dense videos through a diffusion decoder. For video understanding, a dual-flow selection strategy is adopted to extract multi-scale visual representations and integrate supervision on visual tokens. The entire framework is trained in an end-to-end manner.}
  \vspace{-5mm}
  \label{fig:pipeline}
\end{figure*}
\section{Method}
In this section, we introduce the overall framework (Section~\ref{sec:framework}), as illustrated in Figure~\ref{fig:pipeline}. 
Both textual and visual inputs are first converted into discrete tokens and jointly modeled by an autoregressive transformer with a shared codebook. For video generation (Section~\ref{sec:video_gen}), the model predicts sparse keyframe-level visual tokens, which are then decoded into dense videos via a diffusion-based decoder. For video understanding (Section~\ref{sec:video_und}), multi-scale visual representations are extracted using a dual-flow selection strategy, with additional supervision applied at the visual-token level. In the following, we provide a detailed description of each component.

\subsection{Overall Framework}
\label{sec:framework}

\paragraph{Autoregressive Transformer}
Given a multimodal sequence comprising image/video and text inputs, we first discretize these heterogeneous signals into a unified sequence of discrete tokens, enabling a shared representation space across modalities. The unified sequence is modeled autoregressively as:
\begin{equation}
P(x_1, x_2, \cdots, x_T) = \prod_{t=1}^{T} P(x_t \mid x_1, x_2, \cdots, x_{t-1})
\label{eq:autoregressive}
\end{equation}
where the model learns to predict each token conditioned on all its preceding context, regardless of modality.

\vspace{-3mm}
\paragraph{Vision Tokenizer}
The Vision Tokenizer converts images into discrete tokens. Unlike conventional methods like VQ-VAE~\cite{van2017neural, esser2021taming}, which quantize in pixel space, we use TA-Tok~\cite{han2025vision} to tokenize in the semantic feature space, yielding more compact and semantically meaningful representations.

\vspace{-3mm}
\paragraph{Diffusion Decoder} 
The integration of an LLM backbone with an external diffusion module has been extensively explored in the image domain~\cite{dong2023dreamllm,sun2024generative,ge2024seed,pan2025transfer}, where the LLM serves as a prompt-driven embedding enhancer to improve the overall generative quality.

In contrast to previous approaches that autoregressively generate \emph{text-conditioned} signals, our method enables the model to predict \emph{visual-conditioned} signals. Specifically, for video generation, the model autoregressively predicts keyframe-level visual tokens, which are then provided as visual conditions to the diffusion decoder for dense frame rendering.

\subsection{Video Generation}
\label{sec:video_gen}
\paragraph{Autoregressive Video Modeling in the Semantic Space.}
In the autoregressive transformer, we model video sequences autoregressively in the semantic token space rather than the pixel space.
For a given input video, a set of keyframes is uniformly sampled, and each frame is encoded using a visual tokenizer to obtain discrete visual tokens.
Unlike pixel-space representations~\cite{van2017neural}, which are difficult to compress and require a large number of tokens per frame, semantic tokens provide a compact and semantically meaningful representation of visual content. This compactness reduces computational cost and mitigates redundancy across consecutive frames, making the representation well-suited for sequential modeling. Autoregressive training follows a standard next-token prediction objective, where the model learns to predict each semantic token conditioned on all previously observed tokens.
\vspace{-3mm}
\paragraph{Noise-Controlled Visual Conditioning.}
\label{sec:ncvc}
While the autoregressive transformer can effectively predict semantic tokens for keyframes, these sparse representations alone are insufficient for producing temporally dense videos.
Therefore, we employ a diffusion-based decoder to render the predicted semantic tokens into continuous, high-fidelity video sequences.
However, the semantic encoding inevitably discards certain low-level appearance details, which are crucial for video generation tasks such as image-to-video (I2V) generation. To mitigate this issue, we incorporate information from the reference frame as an additional conditioning signal.
Inspired by Diffusion Forcing~\cite{chen2024diffusion}, we control the visual conditioning by applying different noise levels to the latent representations. Specifically, given video latents \( z \in \mathbb{R}^{C \times T \times H \times W} \) encoded by the VAE, we introduce a binary mask \( M \in \{0,1\}^{T} \) to regulate the conditioning process. For text-to-video generation, all mask values are set to 1, meaning that Gaussian noise is added to all frames during training. In contrast, for image-to-video or video-to-video generation, the mask values corresponding to the conditioned frames are set to 0 (no noise added), while the remaining frames follow the standard diffusion noise schedule. Formally, the noisy latent is defined as:
\begin{equation}
\tilde{z}_t = \sqrt{\bar{\alpha}_t} \, z + M \odot \sqrt{1 - \bar{\alpha}_t} \, \epsilon,
\quad \epsilon \sim \mathcal{N}(0, I)
\label{eq:mask_diffusion}
\end{equation}

where \( \bar{\alpha}_t \) denotes the cumulative noise coefficient at timestep \( t \), and \( \odot \) represents element-wise multiplication. This unified masking mechanism ensures consistent training across video generation tasks and enables autoregressive generation, allowing video clips to be seamlessly extended or stitched over time.

\subsection{Video Understanding}
\label{sec:video_und}
For video understanding, our core design leverages \emph{dual-flow selection} for multi-scale visual feature extraction and \emph{vision supervision} to provide auxiliary guidance on vision tokens.

\begin{figure}[t]
  \centering
  \includegraphics[width=\linewidth]{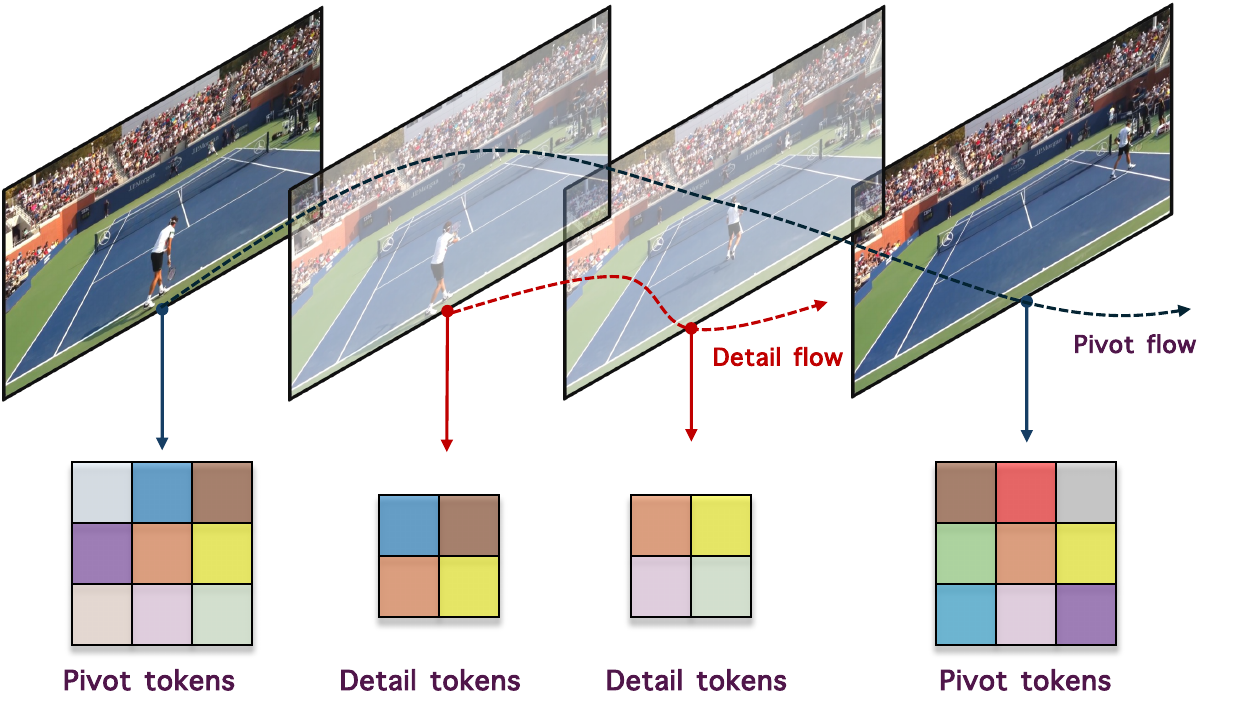} 
  \captionof{figure}{\textbf{Illustration of dual-flow selection.} \textcolor[RGB]{80,22,74}{Pivot} frames are uniformly sampled for main plots; \textcolor[RGB]{192,0,0}{detail} frames are sampled between pivots for fine-grained information and undergo a higher token pooling scale.}
  \label{fig:dualflow}
  \vspace{-3mm}
\end{figure}
\vspace{-3mm}
\paragraph{Dual-flow Selection}
\label{sec:dual_flow}
Video understanding faces the challenge of handling a large number of visual tokens. To address this while avoiding excessive computational cost, we introduce a Dual-flow Selection mechanism.
This approach is based on the observation that videos typically consist of multiple coherent semantic segments, or ``plots''. We treat the initial frame of each segment (the ``pivot'' frame) as containing the highest-level semantic information, while the subsequent ``detail'' frames provide context and fine-grained dynamic information. Accordingly, video frames are processed through two information flows with distinct downsampling scales, as illustrated in Figure~\ref{fig:dualflow}.

\noindent \textbf{Pivot Flow}
First, we uniformly sample $N_p$ frames from the complete video sequence $V = \{f_t\}_{t=1}^T$ to form the set of pivot frames $\mathcal{F}_p$. We hypothesize that these frames $f_p \in \mathcal{F}_p$ contain most of the semantic events. Each frame $f_p$ is converted into a token sequence $X_p = \mathcal{E}(f_p)$ via visual tokenizer $\mathcal{E}$.
Given the semantic importance of pivot frames, we adopt a conservative downsampling strategy to preserve critical information. Specifically, we apply a token average pooling operation $\text{AvgPool}(\cdot, s)$ with a small scale $s_p$:
\begin{equation}
X'_p = \text{AvgPool}(X_p, s_p = 2)
\end{equation}
This smaller pooling scale ensures that the rich spatio-temporal semantic information required for identifying core events is preserved with high fidelity.

\noindent \textbf{Detail Flow}
To further capture the fine-grained dynamics and contextual information within each semantic segment, we uniformly sample $N_d$ frames between each consecutive pair of pivot frames $(f_{t_i}, f_{t_{i+1}})$ (where $f_{t_i}, f_{t_{i+1}} \in \mathcal{F}_p$), forming the detail frame set $\mathcal{F}_d^{(i)} \subset \{f_t | t_i < t < t_{i+1}\}$.

These detail frames $f_d \in \mathcal{F}_d^{(i)}$ are similarly transformed into token sequences $X_d = \mathcal{E}(f_d)$, which provide supplementary semantic information. To efficiently manage the token budget, we apply a more aggressive downsampling strategy using a larger pooling scale $s_d$:
\begin{equation}
    X'_d = \text{AvgPool}(X_d, s_d = 3)
\end{equation}

\noindent \textbf{Mechanism Rationale}
By distinguishing between pivot and detail frames and adopting different pooling strategies with $s_p < s_d$, our dual-flow selection mechanism strategically allocates the model's computational budget. It preserves key information at critical semantic boundaries (pivots) while efficiently compressing supplementary information within semantic segments (details). This strategy enables the model to process a larger total number of frames $\mathcal{F} = \mathcal{F}_p \cup (\bigcup_i \mathcal{F}_d^{(i)})$, achieving a more comprehensive understanding of long videos within a manageable computational cost.

\vspace{-3mm}
\paragraph{Vision Supervision}
\label{sec:vision_supervision}
To further enhance the representational power of the visual tokens extracted by our model, we introduce an auxiliary self-supervised task.
We term this auxiliary task ``Vision Supervision'', drawing inspiration from the masked autoencoding objectives~\cite{he2022masked} commonly used in transformer-based sequence models.

The core idea is to perform masked token prediction on the selected visual-token sequence. For a given video, we concatenate the processed pivot tokens and detail tokens into a sequence $\mathcal{X}' = [\mathcal{F}_p', \mathcal{F}_d']$, where $\mathcal{F}_p' = \{X'_{p,i}\}$ and $\mathcal{F}_d' = \{X'_{d,j}\}$ denote the pooled tokens from the pivot and detail flows, respectively.

During training, we randomly replace a subset of tokens in $\mathcal{X}'$ with mask tokens before feeding the sequence into the model. Let $\mathcal{M}$ be the set of masked positions and $\mathcal{U}$ be the set of unmasked positions. The model is trained to recover the original discrete visual token IDs at the masked positions $\mathcal{X}'_{\text{masked}} = \{X'_k | k \in \mathcal{M}\}$, conditioned on the remaining unmasked tokens $\mathcal{X}'_{\text{unmasked}} = \{X'_k | k \in \mathcal{U}\}$ and the surrounding multimodal context.

Let the model’s predictions for the masked tokens be $\hat{\mathcal{X}}'_{\text{masked}} = \{\hat{X}'_k | k \in \mathcal{M}\}$.
The supervision objective minimizes the cross-entropy loss between the predicted and original token IDs only at the masked positions, encouraging the model to learn context-aware spatio-temporal representations for downstream video understanding.

\begin{table*}[t]
\centering
\caption{\textbf{Comparison with text-to-video and image-to-video models on the VBench and VBench++ benchmarks.}}
    \resizebox{0.85\linewidth}{!}{
\begin{tabular}{lccccccccc}
\toprule
\textbf{Model} & \textbf{\#Params} & \multicolumn{3}{c}{\textbf{VBench Scores} $\uparrow$} & \multicolumn{3}{c}{\textbf{VBench++ Scores} $\uparrow$} \\
\cmidrule(lr){3-5} \cmidrule(lr){6-8}
& & Total Score & Quality Score & Semantic Score & Total Score & Quality Score & I2V Score \\
\midrule
\rowcolor{gray!20}\textit{Diffusion models} & & & & & & & \\
OpenSoraPlan V1.3~\cite{lin2024open} & - & 77.23 & 80.14 & 65.62 & - & - & - \\
Show-1~\cite{zhang2025show} & 6B & 78.93 & 80.42 & 72.98 & - & - & - \\
LTX-Video~\cite{hacohen2024ltx} & 1.9B & 80.15 & 82.52 & 70.68 & - & - & - \\
CogVideoX~\cite{yang2024cogvideox} & 5B & 81.91 & 83.05 & 77.33 & 86.7 & 78.6 & 94.8 \\
Step-Video~\cite{ma2025step} & 30B & 81.83 & 84.46 & 71.28 & 88.4 & 81.2 & 95.5 \\
Gen3~\cite{gen3alpha} & - & 82.32 & 84.11 & 75.17 & - & - & - \\
Kling1.6~\cite{kling16} & - & 83.40 & 85.00 & 76.99 & - & - & - \\
HunyuanVideo~\cite{kong2024hunyuanvideo} & 13B & 83.24 & 85.09 & 75.82 & 86.8 & 78.5 & 95.1 \\
Wan2.1~\cite{wan2025wan} & 14B & 84.70 & 85.64 & 80.95 & 86.9 & 80.8 & 92.9 \\
\midrule
\rowcolor{gray!20}\textit{Autoregressive models} & & & & & & & \\
VILA-U~\cite{wu2024vila} & 7B & 74.01 & 76.26 & 65.04 & - & - & - \\
Lumos-1~\cite{yuan2025lumos} & 3.6B & 78.32 & 79.52 & 73.51 & 84.7 & 76.1 & 93.3 \\
NOVA~\cite{deng2024autoregressive} & 0.6B & 80.12 & 80.39 & 79.05 & - & - & - \\
CausVid~\cite{yin2025slow} & 1.3B & 81.46 & 84.05 & 69.80 & - & - & - \\
Pyramid Flow~\cite{jin2024pyramidal} & 2B & 81.72 & 84.74 & 69.62 & - & - & - \\
VideoMAR~\cite{yu2025videomar} & 1.4B & - & - & - & 84.8 & 75.6 & 94.0 \\
\midrule
\rowcolor{gray!20}\textit{Unified models} & & & & & & & \\
HaploOmni~\cite{xiao2025haploomni} & 9B & 78.10 & - & - & - & - & - \\
Emu3~\cite{wang2024emu3} & 8B & 80.96 & \textbf{84.09} & 68.43 & - & - & - \\
Show-o2~\cite{xie2025show} & 2B & 81.34 & 82.10 & 78.31 & 85.9 & 77.7 & 94.1 \\
\textbf{\methodname{} (Ours)} & 3B & \textbf{82.58} & 82.64 & \textbf{82.35} & \textbf{86.7} & \textbf{79.4}  & \textbf{94.1} \\
\bottomrule
\end{tabular}}
\label{tab:vbench}
\vspace{-3mm}
\end{table*}
\section{Experiment}

\subsection{Experiment Setup}
\noindent \textbf{Dataset}
The training is divided into two stages. The first stage involves joint training on image-text modalities, while the second stage fine-tunes the model for downstream video tasks. The first-stage dataset comprises approximately 100 million text-image samples, including Blip3o~\cite{chen2025blip3}, JourneyDB~\cite{sun2023journeydb}, DenseFusion~\cite{li2024densefusion}, FineVision~\cite{wiedmann2025finevision}, LLaVA-OneVision~\cite{li2024llava}, and Infinity-mm~\cite{gu2024infinity}.
The second-stage dataset comprises 5 million video samples sourced from Koala-36M~\cite{wang2025koala}, OpenVid-1M~\cite{nan2024openvid}, LLaVA-Video-178K~\cite{zhang2024video}, together with image data from ShareGPT-4o-Image~\cite{chen2025sharegpt}, Blip3o-60k~\cite{chen2025blip3} and Echo-4o~\cite{ye2025echo}.

\noindent \textbf{Benchmarks}
We primarily evaluate our model on both video generation and video understanding benchmarks. For video generation, we use the VBench series~\cite{huang2024vbench, huang2024vbench++} benchmarks to assess the quality and fidelity of generated videos.
For video understanding, we evaluate performance on established multimodal understanding benchmarks, including VideoMME~\cite{fu2025video}, LongVideo Bench~\cite{wu2024longvideobench}, Egoschema~\cite{mangalam2023egoschema}, Next-QA~\cite{xiao2021next} and MLVU~\cite{zhou2025mlvu}.

\noindent \textbf{Training Details}
We employ the Qwen2.5-3B model~\cite{yang2025qwen3} as the base autoregressive transformer, while the diffusion decoder is initialized from Wan2.1-1.3B~\cite{wan2025wan}. The vision tokenizer is adopted from Tar~\cite{han2025vision}, yielding a total of 65,536 discrete visual tokens. A lightweight two-layer MLP is further introduced to bridge the hidden states of the autoregressive transformer with the diffusion decoder. 
Training is conducted in two stages. The first stage involves large-scale image-text autoregressive pretraining to establish strong multimodal alignment. For Stage 1 training, we use a batch size of 256 and train the model for 200,000 steps.
For video generation training, we train the AR and diffusion modules jointly with a batch size of 64 for 60,000 steps at a resolution of 49$\times$832$\times$480. The videos are sampled at 8 FPS, with keyframes extracted every 2 seconds.
To enhance semantic understanding, video data are mixed with text-to-image (T2I) samples at a 2:1 ratio during training. Additionally, during video training, there is a 30\% probability of conditioning on the first frame to train image-to-video (I2V) generation.
For ablation studies, we used a resolution of 33$\times$832$\times$480 and trained the model for 20,000 steps for faster evaluation.
For video understanding training, we set batch size as 16 and train the model for 80,000 steps, with 30 frames sampled from each video in total, including 10 pivot frames and 20 detail frames.

\begin{table*}[ht]
\centering
\small
\setlength{\tabcolsep}{4pt}
\renewcommand{\arraystretch}{1.15}
\caption{\textbf{Comparison of video understanding models across multiple benchmarks.} }
    \resizebox{0.85\linewidth}{!}{
\begin{tabular}{lc|ccccc}
\toprule
\textbf{Model} & \textbf{\#Params.}  &
\textbf{VideoMME$\uparrow$} &
\textbf{MLVU$\uparrow$} &
\textbf{LongVideoBench$\uparrow$} &
\textbf{NextQA$\uparrow$} &
\textbf{EgoSchema$\uparrow$} \\
\midrule
\rowcolor{gray!20}\multicolumn{7}{l}{\textit{Proprietary Und. Only Models}} \\
GPT-4V~\cite{openai2023gpt4v} & -  
& 57.0 & 43.5 & 61.3 & - & 55.6  \\
GPT-4o~\cite{hurst2024gpt} & - 
& 71.9  & 64.6 & 66.7 & -  & 72.2 \\
Gemini-1.5-Flash~\cite{team2024gemini} & -  
& 70.3 & 51.3 &  61.6  & - & - \\
Gemini-1.5-Pro~\cite{team2024gemini} & - 
& 75.0 & - & 64.0 & - & 63.2 \\
\midrule
\rowcolor{gray!20}\multicolumn{7}{l}{\textit{Open-source Und. Only Models}} \\
LongVA~\cite{zhang2024long} & 7B &  52.6 & - & - & 68.3& -  \\
VideoLLaMA2~\cite{cheng2024videollama} & 7B 
&  47.9 & - & 58.2 & - & 51.7 \\
VideoLLaMA3~\cite{zhang2025videollama} & 7B 
&  66.2 & 73.0 & 63.3 & 84.5 & 63.3 \\
LLaVA-OV~\cite{li2024llava} & 7B 
&  58.2 & 56.3 & 56.4 & 79.4 & 60.1 \\
LLaVA-Video~\cite{zhang2025llavavideo} & 7B 
& 63.3 & 70.8 & 58.2 & 83.2 & 57.3 \\
\midrule
\rowcolor{gray!20}\multicolumn{7}{l}{\textit{Unified Multimodal Models}} \\
HaploOmni~\cite{xiao2025haploomni} & 9B 
& - &- & - & - & 47.1 \\
Show-o2~\cite{xie2025show} & 7B 
& 57.4 & - & 55.5 & \textbf{79.0} & -  \\
\textbf{\methodname{} (Ours)} & 3B 
& \textbf{69.4} & \textbf{71.6} & \textbf{60.1} & 76.3 &  \textbf{52.9}  \\
\bottomrule
\end{tabular}}
\label{tab:video_und}
\vspace{-2mm}
\end{table*}
\noindent \textbf{Inference Details}
For text-to-video generation, we use the original prompt suite provided in VBench, which contains 946 prompts across 16 dimensions, without applying prompt rewriting.
For evaluating video understanding performance, we adopt LMMs-Eval~\cite{zhang2024lmmsevalrealitycheckevaluation}.

For video generation inference, we configure the autoregressive transformer with top-k = 256 and top-p = 0.95 for visual token sampling, while the diffusion decoder operates with a classifier-free guidance (CFG) scale of 4.0 and performs 50 denoising steps to generate the final video frames.

For video understanding, we evaluate VideoMME without subtitles, LongVideoBench on the validation set, and EgoSchema on the test set. The MLVU accuracy is reported as the mean over all subtasks. We use the same frame selection mechanism from the training stage, with 30 frames sampled in total, 10 and 20 for pivot and detail frames.

\subsection{Main Results}
\paragraph{Video Generation}
We evaluate text-to-video performance on VBench~\cite{huang2024vbench} and image-to-video performance on VBench++\cite{huang2024vbench++}. As shown in Table~\ref{tab:vbench}, \methodname{} achieves leading performance among unified video understanding-generation models, while showing competitive results against specialized video generation models. Furthermore, its strong image-to-video results demonstrate the framework’s versatility for general-purpose video generation.

\vspace{-5mm}
\paragraph{Video Understanding}
As shown in Table~\ref{tab:video_und}, we evaluate our model across a range of video understanding benchmarks. Our approach achieves competitive performance compared with strong open-source understanding-only video models.
Compared with prior unified approaches, our model shows clear gains on VideoMME and LongVideoBench, achieving 69.4 and 60.1, respectively.
It is worth noting that competing models, both understanding-only and unified, typically contain over 7B parameters, whereas our 3B-parameter model achieves comparable or superior performance. This underscores the strong parameter efficiency and competitive capability of our architecture.

\subsection{Ablation Studies}
In this section, we provide an in-depth analysis of the proposed method.

\vspace{-3mm}
\paragraph{Visual Condition vs. Text Condition}
In Table~\ref{tab:diffusion_cond}, we investigate the benefits of employing visual tokens as conditioning inputs in place of text tokens.
The results show that conditioning on visual tokens consistently yields better generation quality compared to text-based conditioning, suggesting that visual representations provide richer and more fine-grained context for guiding video generation. Interestingly, when visual and textual tokens are used jointly, the performance is comparable to or even slightly lower than using visual tokens alone. This finding further highlights that visual tokens alone provide sufficiently expressive conditioning signals for effective modeling, revealing that text embeddings may not serve as the optimal conditioning modality for video generation.

\begin{table}[t]
  \centering
  \caption{\textbf{Ablation study on diffusion decoder conditioning.}}
    \resizebox{\linewidth}{!}{%
  \begin{tabular}{c c | c c c}
    \toprule
    vision cond. & text cond. & \multicolumn{3}{c}{\textbf{VBench Scores} $\uparrow$} \\
    \cmidrule(lr){3-5}
    & & Total Score & Quality Score & Semantic Score \\
    \midrule
    \checkmark &  &\textbf{82.13} & \textbf{83.36} & \textbf{77.21} \\
     & \checkmark & 78.61 & 81.70 & 66.28\\
     \checkmark & \checkmark & 81.43 & 82.57 & 76.86\\ 
    \bottomrule
  \end{tabular}
  }
  \label{tab:diffusion_cond}
  \vspace{-2mm}
\end{table}

\begin{table}[t]
  \centering
  \caption{\textbf{Analysis of Visual Token Numbers.}}
      \resizebox{\linewidth}{!}{%
  \begin{tabular}{c  c c c | c}
    \toprule
    Keyframes & Tokens/Frame & Total & Keyframes Interval & \textbf{VBench Score} $\uparrow$ \\
    \midrule
    3 & 81 & 243 & 2.0 s & 81.78 \\
    5 & 81 & 405 & 1.0 s & 81.43 \\
    9 & 81 & 729 & 0.5 s & 81.14 \\
    \midrule
    2 & 169 & 338 & 4.0 s & 81.31 \\
    3 & 169 & 507 & 2.0 s & \textbf{82.13}  \\
    5 & 169 & 845 & 1.0 s & 81.72 \\
    9 & 169 & 1521 & 0.5 s & 81.46 \\
    \bottomrule
  \end{tabular}}
  \label{tab:ablation_vision_numbers}
  \vspace{-3mm}
\end{table}
\vspace{-3mm}
\paragraph{Analysis of Visual Token Numbers}
Table~\ref{tab:ablation_vision_numbers} analyzes the impact of varying the number of keyframes and per-frame visual tokens on performance. The results indicate that more visual tokens do not necessarily lead to better performance; we find that predicting keyframes at 2s intervals yields the best results, and moderately increasing the number of tokens per frame can also bring performance gains.

\begin{figure*}[t]
  \centering
     \includegraphics[width=\linewidth]{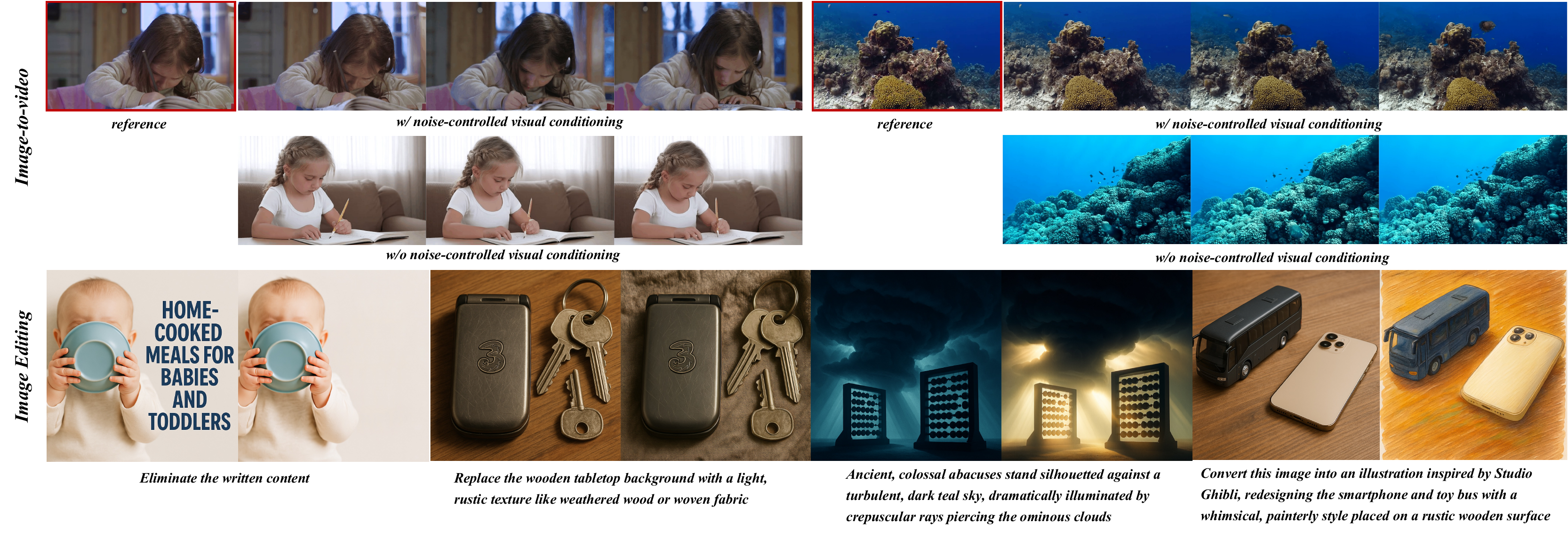} 
    \vspace{-6mm}
  \caption{\textbf{Effectiveness of noise-controlled visual conditioning.} Since the autoregressive module predicts semantic visual tokens, it primarily ensures semantic consistency across frames. Incorporating noise-controlled visual conditioning further enhances fine-grained visual coherence while maintaining semantic alignment. This modeling approach is also applicable to image editing tasks, treating them as special two-frame video sequences.}
  \label{fig:visual_demo}
  \vspace{-3mm}
\end{figure*}

\vspace{-3mm}
\paragraph{Effectiveness of Dual-flow Selection}
Table~\ref{tab:ablation_video_und} demonstrates that the proposed dual-flow selection significantly enhances temporal modeling. Compared to the baseline, which uniformly samples 20 frames with a single pooling scale of 2, the dual-flow design (Section~\ref{sec:dual_flow}) captures richer temporal cues while avoiding excessive visual tokens, resulting in consistent improvements across all benchmarks.

\begin{table}[t]
  \centering
  \caption{\textbf{Ablation Study on Video Understanding.}}
    \resizebox{\linewidth}{!}{%
  \begin{tabular}{c c | c c c c}
    \toprule
    dual-flow & gen. sup.
 & {\textbf{VideoMME} $\uparrow$} & {\textbf{MLVU} $\uparrow$} & {\textbf{LongVideoBench}$\uparrow$} & {\textbf{NextQA} $\uparrow$} \\
    \midrule
      &  & 60.04 & 64.78 & 53.83 & 49.46 \\
     \checkmark & & 60.48 & 65.64 & 54.32 & 58.84\\
     \checkmark & \checkmark & \textbf{66.89} & \textbf{66.46} & \textbf{58.74} & \textbf{76.41}\\ 
    \bottomrule
  \end{tabular}
  }
  \label{tab:ablation_video_und}
  \vspace{-3mm}
\end{table}
\vspace{-3mm}
\paragraph{Generative Supervision for Understanding}
Building on the dual-flow setup, we add the generative supervision loss (Section~\ref{sec:vision_supervision}). This objective encourages the model to predict visual tokens from surrounding context, providing an auxiliary training signal in addition to text-only supervision.
As shown in Table~\ref{tab:ablation_video_und}, this additional supervision leads to further improvements beyond dual-flow, suggesting that visual-token prediction can complement standard video understanding training.
We hypothesize that these gains may come from the dense supervision provided by autoregressive token prediction over video sequences, which encourages the model to encode temporal and semantic structure more explicitly.

\subsection{Qualitative Analysis}
This section provides qualitative results on image-to-video and image editing tasks, as shown in Figure~\ref{fig:visual_demo}.

\vspace{-3mm}
\paragraph{Effectiveness of Noise-controlled Visual Conditioning}
Although using semantic visual tokens significantly reduces the number of tokens, projecting them into the semantic space inevitably leads to the loss of fine-grained details. This can cause inconsistency between the generated video and the first-frame condition in tasks such as image-to-video (I2V) generation. To alleviate this issue, we propose noise-controlled visual conditioning (Section~\ref{sec:ncvc}), which effectively mitigates such inconsistencies and yields visually consistent results, as demonstrated in Figure~\ref{fig:visual_demo}.

\vspace{-3mm}
\paragraph{Reframing Image Editing as Video Generation}
Within our unified framework, image editing can be naturally interpreted as a form of dynamic video modeling. Conventional image editing takes a reference image and a text prompt to produce a target image. In our formulation, we treat the reference–target pair as a two-frame video sequence, allowing the model to capture the transition dynamics between the original and edited states. As illustrated in Figure~\ref{fig:visual_demo}, our method generates coherent and faithful editing results, demonstrating the compatibility and unifying potential of our video-centric modeling paradigm.




\section{Conclusion}
In this paper, we introduce \methodname{}, a unified framework for video generation and understanding. By transforming multimodal signals into discrete tokens, our method establishes a unified representation space for different modalities.
This enables coherent modeling of visual semantics and temporal dynamics through a hybrid architecture that integrates autoregressive semantic prediction with diffusion-based rendering. Extensive experiments on both generation and understanding benchmarks demonstrate the effectiveness of our approach and show that learning from visual tokens can benefit both tasks: offering richer conditioning for generation and providing auxiliary generative supervision for understanding. Overall, this work underscores the promise of video-centric modeling as a foundation for advancing unified multimodal learning.

\vspace{-5mm}
\paragraph{Limitations}
Due to computational constraints, the current data scale and model size for video generation and understanding remain limited; thus, our framework represents an initial step in this direction. We believe it is inherently well-suited for multimodal reinforcement learning and in-context learning, which we plan to explore further in future work.

{
    \small
    \bibliographystyle{ieeenat_fullname}
    \bibliography{main}
}

\clearpage
\setcounter{page}{1}
\maketitlesupplementary

\appendix

\section{Qualitative Results}
\label{sec:vis}
This section provides additional examples that further demonstrate the model’s capabilities in both video understanding and video generation.

\subsection{Video Generation}
\begin{figure*}[b]
    \centering
    \includegraphics[width=1\linewidth]{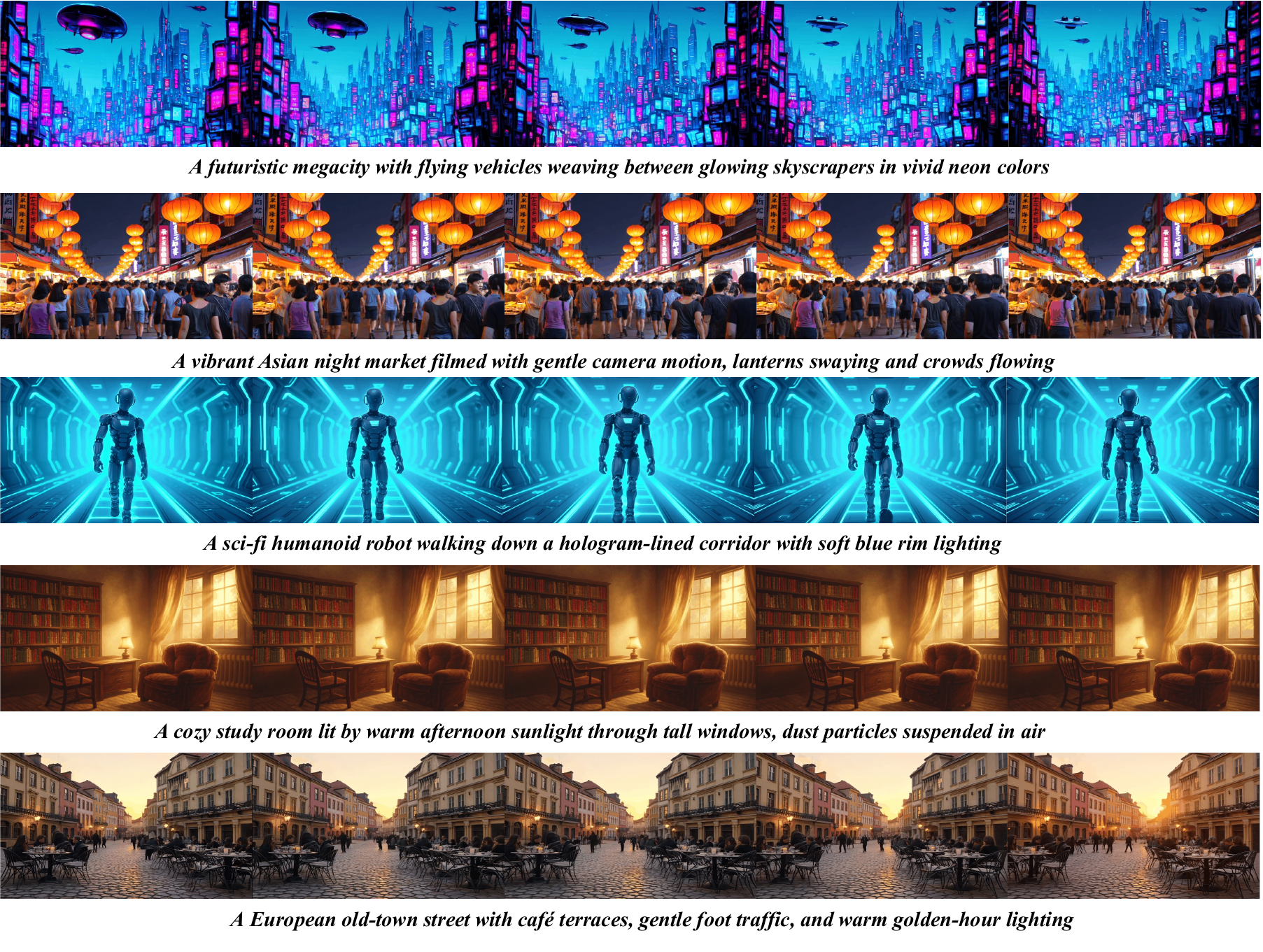}
    \caption{\textbf{Visualizations of text-to-video generation.}}
    \label{fig:t2v_demo}
\end{figure*}

As shown in Figure~\ref{fig:t2v_demo}, our model is capable of generating a diverse set of visually appealing video sequences from textual descriptions, while demonstrating strong instruction-following capabilities.

\subsection{Video Understanding}
\begin{figure*}[htb]
    \centering
    \includegraphics[width=1\linewidth]{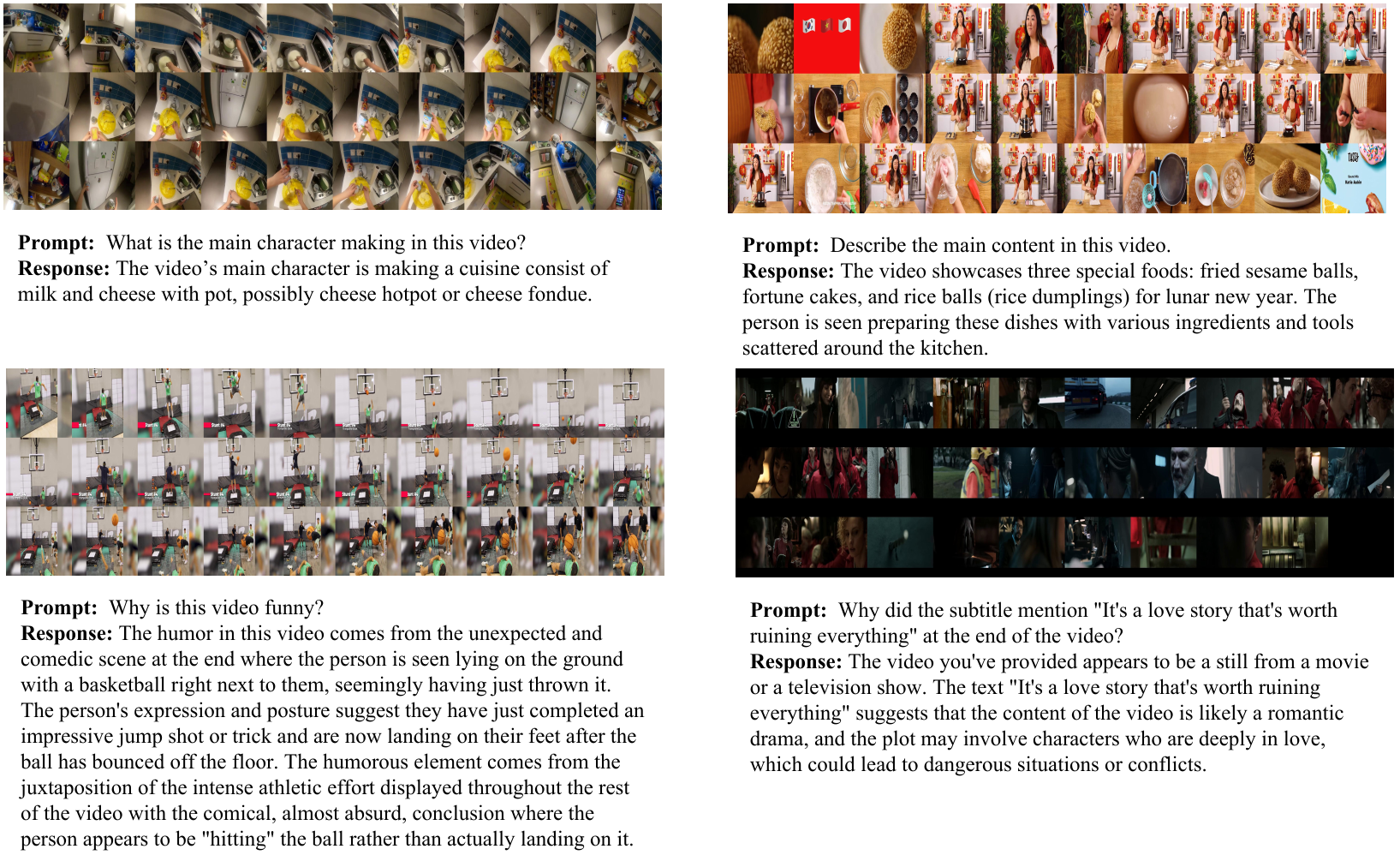}
    \caption{\textbf{Video understanding demos illustrating the model's ability to extract and analyze video content.}}
    \label{fig:v_demo}
\end{figure*}

Here we present additional qualitative examples showcasing the capabilities of Vega in responding to various video inputs and corresponding text prompts. Figure~\ref{fig:v_demo} provides four examples demonstrating the video understanding capability, including (a) an example illustrating the model's ability to infer fine-grained content with basic knowledge; (b) an example illustrating the model's ability to understand hours-long videos; (c) an example illustrating the model's ability to perform causal reasoning, such as explaining humorous content in the video; and (d) an example illustrating the model's ability to understand the logical relationship between two long-duration plots.

\section{Implementation Details}
\paragraph{Training Format}
The first-stage model is trained on image–text pairs, and to ensure compatibility with video inputs, the specific training strategy and data format are described as follows.
For video understanding tasks, we read videos as byte streams and integrate them with the corresponding text conversations, forming the standard conversation format used by OpenAI.

\begin{verbatim}
    {
      "video": video_byte_stream,
      "conversations":  
      [{'from': 'human', 
      'value': <video>\n+Question}
     {'from': 'gpt', 
     'value': Answer}]
    }
\end{verbatim}

Similarly, for video generation tasks, the training format is organized as shown below.

\begin{verbatim}
    {
      "video": video_byte_stream,
      "conversations":  
      [{'from': 'human', 
      'value': text prompt}
     {'from': 'gpt', 
     'value': <video>}]
    }
\end{verbatim}

During both training and inference, we read the video as a byte stream and select the corresponding frames according to the designated sampling strategy. The $<video>$ segment is then replaced with the corresponding visual tokens.

\paragraph{Evaluation Details}
We conduct evaluations on video understanding benchmarks using the open-source project lmms-eval. For the VideoMME and MLVU benchmarks, we directly report the scores from lmms-eval. However, on the multiple-choice tasks of the NextQA, LongVideoBench, and EgoSchema benchmarks, the model outputs explanatory content beyond the options. This caused the lmms-eval filter to fail parsing the answers. To address this, we implement a custom filter that extracts the first character of the response text as the model's answer. We then compare this answer with the ground truth to calculate the scores.

\paragraph{Failure Analysis}
Here, we provide representative failure cases in video understanding and analyze the key factors that prevent our model from correctly handling these scenarios.
\begin{figure*}[htb]
    \centering
    \includegraphics[width=1\linewidth]{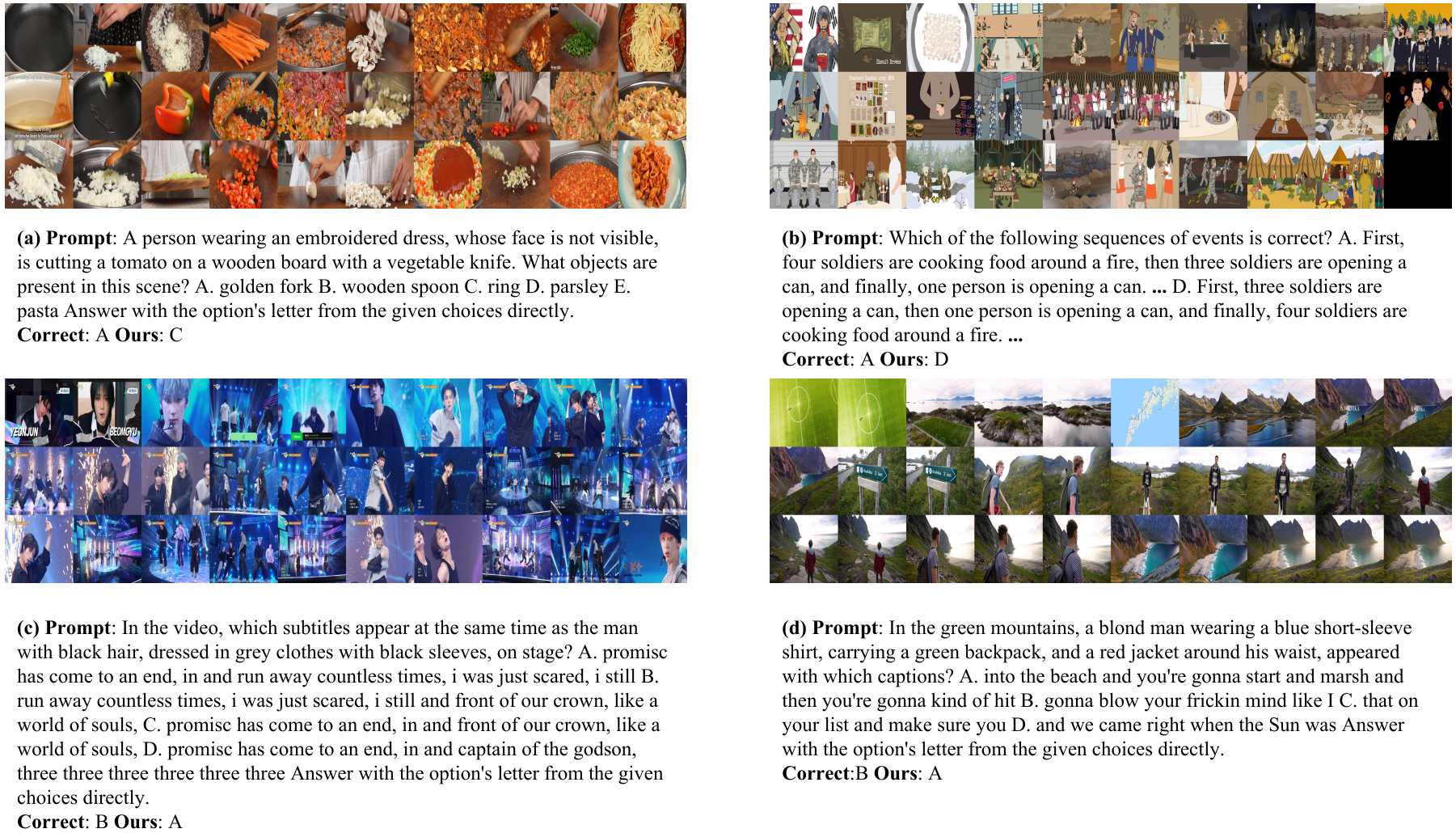}
    \caption{\textbf{Failure cases of our proposed model, on video understanding task.}}
    \label{fig:failure}
\end{figure*}
The observed cases demonstrate the model’s limitations in fine-grained frame selection and content understanding. Given the high frame count of long videos, sampling only 30 frames is insufficient to capture all critical information, leading to performance gaps on detail-oriented benchmarks. Furthermore, the use of token pooling compromises the model’s ability to perceive fine-grained spatial details, including small visual elements such as subtitles. A detailed analysis of representative cases is provided below:
In case (a), the model mistakenly identifies the sliced chilis as tomatoes, leading it to conclude that the ring on the chef’s hand appears in the scene. In reality, the chef is slicing tomatoes with a golden fork, but the relevant frames are not included in the sampled inputs.
In case (b), the model fails to correctly interpret the frame in which the soldier is opening a can, resulting in an incorrect ordering of the scenes. (As the original prompt is too long to display, for simplicity we only show the right and wrong answers chosen by our method.)
In cases (c) and (d), the model is unable to capture the subtitle or caption information present in the required frames, and thus cannot produce the correct answers.

\end{document}